\documentclass[review]{elsarticle}

\usepackage{lineno,hyperref}
\usepackage{graphicx}
\usepackage{comment}
\usepackage{amsmath,amssymb}
\usepackage{color}
\usepackage{url}
\usepackage{float}
\usepackage{subfig}
\usepackage{booktabs}
\usepackage{multirow}
\usepackage{multicol}
\usepackage{adjustbox}
\usepackage[linesnumbered,ruled,vlined]{algorithm2e}
\DontPrintSemicolon
\SetKwInput{KwInput}{Input}                
\SetKwInput{KwOutput}{Output} 
\SetKwProg{Init}{init}

\modulolinenumbers[5]
\DeclareMathOperator*{\argmax}{arg\,max}

\journal{arxiv}

\begin{document}

\begin{frontmatter}

\title{Efficient Curriculum based Continual Learning with Informative Subset Selection for Remote Sensing Scene Classification}

\author[ee]{S Divakar Bhat\corref{mycorrespondingauthor}}
\author[csre]{Biplab Banerjee}

\author[ee]{Subhasis Chaudhuri}
\author[csre]{Avik Bhattacharya}

\address[ee]{Department
of Electrical Engineering, Indian Institute of Technology Bombay,
Mumbai, Maharashtra 400076 India.}
\address[csre]{Centre of Studies in Resources Engineering, Indian Institute of Technology Bombay, Mumbai, Maharashtra 400076 India.}

\begin{abstract}
We tackle the problem of class incremental learning (CIL) in the realm of land-cover classification from optical remote sensing (RS) images in this paper. The paradigm of CIL has recently gained much prominence given the fact that data are generally obtained in a sequential manner for real-world phenomenon. However, CIL has not been extensively considered yet in the domain of RS irrespective of the fact that the satellites tend to discover new classes at different geographical locations temporally. With this motivation, we propose a novel CIL framework inspired by the recent success of replay-memory based approaches and tackling two of their shortcomings. In order to reduce the effect of catastrophic forgetting of the old classes when a new stream arrives, we learn a curriculum of the new classes based on their similarity with the old classes. This is found to limit the degree of forgetting substantially. Next while constructing the replay memory, instead of randomly selecting samples from the old streams, we propose a sample selection strategy which ensures the selection of highly confident samples so as to reduce the effects of noise. We observe a sharp improvement in the CIL performance with the proposed components. Experimental results on the benchmark NWPU-RESISC45, PatternNet, and EuroSAT datasets confirm that our method offers improved stability-plasticity trade-off than the literature.
\end{abstract}

\end{frontmatter}


\section{Introduction}
Remote sensing image analysis has been witnessing a surge in the development of innovative solutions to many advanced problems with considerable momentum in the recent past. Reinforcing the notion of data-driven feature learning, deep convolutional neural networks can be attributed the credit for a significant portion of this success. This, along with the large number of satellites being deployed with its highly advanced imaging technologies capable of accumulating abundant data~\cite{BELWARD2015115} has resulted in a peak in the research happening in the field. 
\begin{figure}[h]
\centering
\includegraphics[width=0.7\textwidth]{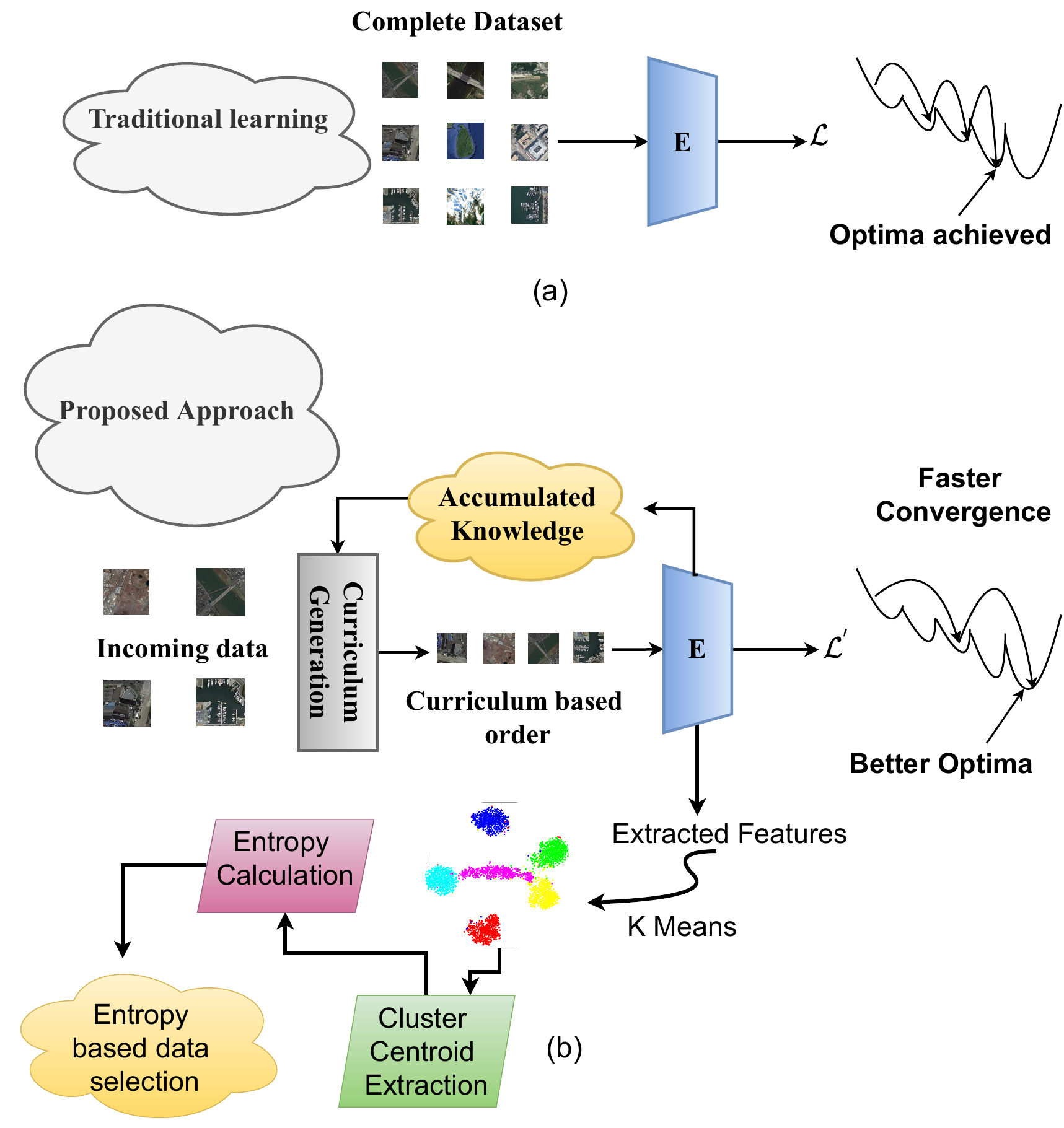}
\caption{Comparison of the conventional learning method and the proposed continual learning scheme. (a) shows a typical deep learning scenario in which the entire dataset is supposed to be available during training. (b) depicts the proposed method for learning incrementally by combining a curriculum learning strategy with continuous learning integrated with the informative subset selection technique.}
\label{fig:motivation}
\end{figure}
Unlike the conventional setting, where the entire information is present during the training phase, the land cover classes are acquired sequentially by the earth observation sensors. Therefore, the data thus collected is prone to limitations in temporal, spatial, and spectral resolution. For instance, a satellite will continuously send raw image data from across the world, creating a rich reservoir of information that could be used for many applications. However, this requires these samples to be annotated meticulously and adequately. The gradual accumulation of these vast amounts of multi-temporal data resulting from continuous inflow from the sensors thus demands a framework to learn incrementally.

However, it is observed that training a deep learning framework on dynamically varying streams of incoming data is particularly complex and demanding. This is due to the fact that while the network tends to forget previously acquired information as a result of its sensitivity to quick updates, it also shows drastically poor reactivity to slow and gradual updates~\cite{gepperth2016incremental}. This is the well-known phenomenon of the stability-plasticity dilemma~\cite{mermillod2013stability}, which is a significant constraint for developing real-time artificial learning systems.
This dramatic decrease in the performance of the model under consideration when it is trained over a continually growing dynamic distribution of data is termed as catastrophic forgetting~\cite{french1999catastrophic}.
The most naive approach that can be adopted to solve this issue is to train the network from scratch every time a new set of data is introduced. However, as expected, this will account for a massive overhead of retaining the whole set of data acquired continually by earth observation sensors from across the globe. Therefore, it is necessary to develop novel learning paradigms that can handle catastrophic forgetting, thus facilitating on the go training of the deep neural network.

As discussed before, despite the advancements made in feature-based learning in remote sensing, the perpetual flow of data from satellites pose severe challenges in terms of memory resource requirements, the need for proper annotation of this vast collection of image data, etc. for making the most out of these rich satellite image data reservoirs. Moreover, not much work has been done in this direction in remote sensing despite the clear possibilities to explore. Also, it is not practically feasible to always expect the availability of having all the class information beforehand. And contrary to the expectation from any efficient learning systems, none of the current methods exploit the similarity between the incoming stream of data with that of the previously acquired information. Furthermore, existing methodologies that employ methods like the rehearsal-based approach do not consider how the selection of old samples could affect the extent of interference the network is subjected to. Memory replay-based methods explicitly hold a steady ground as most suitable for continual learning in satellite images, mainly due to the presence of rich semantic concepts in satellite images that a GAN-based method will fail to generate precisely.

To address these problems, we propose a novel approach to continual learning by employing curriculum learning and an intelligent selection of informative samples for rehearsal-based learning. Introduced in~\cite{bengio2009curriculum}, curriculum learning is an efficient learning paradigm that proposes to present data to the model in a more meaningful manner based on the complexity of the concepts. This is reinforced by~\cite{skinner1958reinforcement,krueger2009flexible} which points out that when a task is decomposed into sub-tasks from easy to complex, animals tend to learn faster. 
We also employ an entropy-based efficient selection of informative samples inspired from works in active learning like~\cite{paul2016efficient}, thus aiming to make the most out of the available budget for sample retention. Commonly used techniques for sample selection for computer vision problems is discussed in~\cite{settles2009active} and have been successfully implemented for tracking~\cite{vondrick2011video}, object detection~\cite{vijayanarasimhan2014large}, etc. 
In this work, we propose a novel entropy-based efficient selection of representative samples for replay coupled with a curriculum-based continual learning approach for remote sensing scene classification. Unlike existing replay-based works in vision like ~\cite{rebuffi2017icarl} and remote sensing works like~\cite{ammour2020continual}, we propose an end-to-end model and do not use any extra network for sample selection. The proposed work consists of the following contributions:
\begin{itemize}
    \item We tackle the problem of CIL in the context of land-cover classification from optical remote sensing data. 
    \item Our method is developed on the premise of rehearsal-based memory replay. However, we propose two novel components within the framework to better tackle the issues related to stability and plasticity. A novel continual learning approach for remote sensing image classification using a curriculum learning technique. We also present a pseudo-teacher-student-based approach for continual learning using the curriculum.

    \item We perform thorough experiments with three optical datasets to demonstrate how a combination of proposed efficient sample selection technique and curriculum learning improves the performance of a continual learning network for satellite image classification.
\end{itemize}
We have provided a brief overview of the proposed approach in the Figure~\ref{fig:motivation} motivating the relevance of the work for real time applications like remote sensing.
\begin{figure*}[t]
\centering
\includegraphics[width=\textwidth]{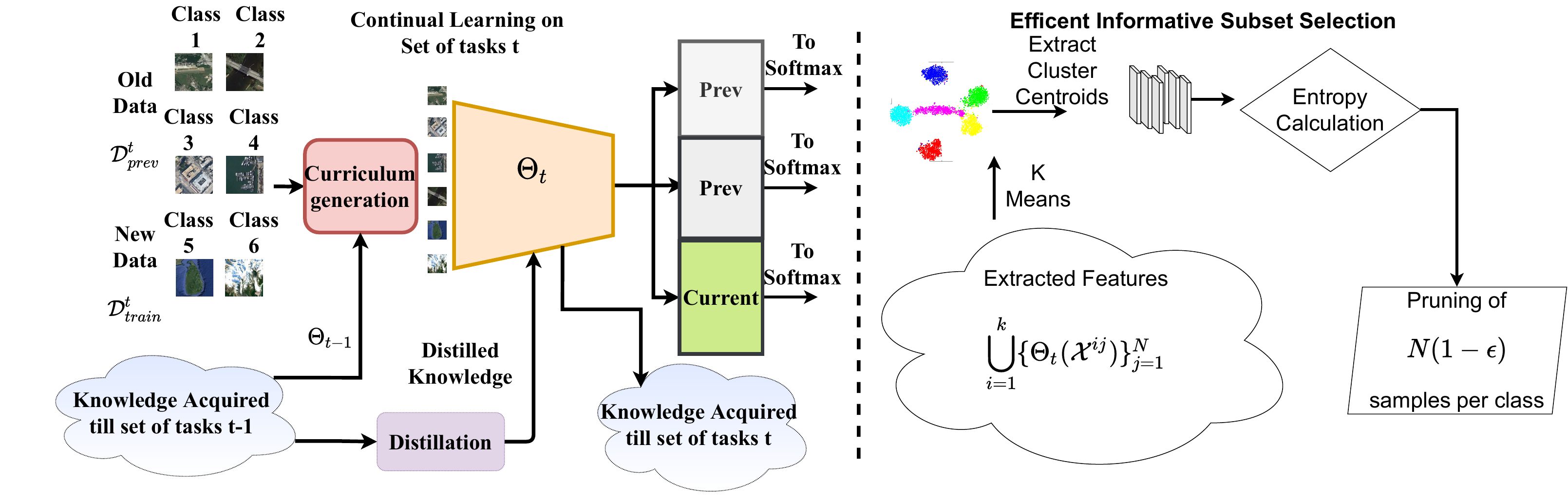} 
\caption{\textbf{The complete pipeline of the proposed framework}. The incoming novel classes are organised into a curriculum for the collection of tasks $t$. Based on the curriculum produced, the network $\Theta_t$ learns the features from a new stream of data. To transfer previously learned information, we rely on knowledge distillation from $\Theta_{t-1}$.}
\label{fig:method}
\end{figure*}
\section{Related Works}
There have been many approaches that has been proposed to alleviate the issue of catastrophic forgetting both using classical and modern methods in feature based learning. Notable of these works are~\cite{10.1007/978-3-540-71701-0_101,bruzzone1999incremental} which proposed to use SVM and RBF based models respectively for training in incremental setting. While, in~\cite{lakshminarayanan2014mondrian}, it was proposed to use hierarchically growing  random forest-based approach.

In~\cite{kirkpatrick2017overcoming} and~\cite{liu2018rotate}, the authors employ an elastic weight consolidation term which was used to assign importance to neurons corresponding both old and new tasks and impose a quadratic penalty on the change between previous and current network. This importance weight computation was made online in~\cite{zenke2017continual}. While the authors align the predictions based on the current task in~\cite{li2017learning}. 

The approach used in~\cite{rebuffi2017icarl} suggests to train at a given instant by using a small fraction of retained samples from previous class along with the new samples. While~\cite{li2017learning} proposes to use knowledge distillation~\cite{hinton2015distilling} for facilitating proper transfer of knowledge from the previous tasks to to the updated network. Whereas in~\cite{castro2018end,tasar2019incremental} recommends an approach where most representative samples are used in combination with the knowledge distillation based approach to mitigate forgetting. In~\cite{wu2018incremental} the authors propose a generative approach where the samples corresponding to old classes are synthesised. 
Most recent works in the field like in~\cite{NIPS2019_9357} the authors propose a replay based approach where only the most conflicted samples are retrieved. While, performance improvement is observed in~\cite{prabhu2020greedy} by greedily storing previous samples and retraining on these during testing. Whereas an expansion-based approach is employed in~\cite{lee2020neural} which is built on the principles of Bayesian nonparametric.

There have been only a handful of works that explored continual learning in the field of remote sensing. These few existing techniques like that in~\cite{ammour2020continual} relies on a separate module for the retrieval of samples for replay. While the authors explore incremental learning for the task of semantic segmentation in~\cite{tasar2019incremental} with only relatively smaller number of new classes per stream. In~\cite{yang2020geoboost}, continual learning is employed  for end-to-end semantic segmentation for global mapping of buildings from VHR satellite images. Whereas the authors have introduced a large scale remote sensing benchmark dataset for continual learning in~\cite{li2020clrs}. 
Despite many works that explore continual learning, none investigate the efficacy of a combination of curriculum learning and informative subset selection for better performance. In this perspective, we propose a versatile framework for continual learning in remote sensing scene classification. 
\section{Methodology}
The complete overview of the proposed framework is illustrated in the Figure~\ref{fig:method}, where it consists of mainly three core modules, a curriculum based pseudo-teacher-student network, a curriculum generator and a informative subset selector. In this section we first formulate the problem definition, and then introduce each section of the framework in detail, respectively.
\subsection{Problem Definition and Notations}
Consider a model $\Theta_{t-1}$ which is to be trained with an incoming set of tasks indexed as $t$ with $k$ novel classes per stream which can be denoted as $\mathcal{D}_{train}^t=\bigcup_{i=1}^k \{ (\mathcal{X}_{(t-1)k + i}^j,\mathcal{Y}_{(t-1)k + i}^j) \}_{j=1}^N$, with $N$ number of samples per class in the given incoming set of tasks. The ordered pair $(\mathcal{X}_{(t-1)k + i}^j,\mathcal{Y}_{(t-1)k + i}^j)$ the $j^{th}$ represents the sample and corresponding ground truth for the $i^{th}$ class in set of task $t$. Our objective is to train the model on the new stream of data without forgetting the previously acquired information resulting in the updated model $\Theta_t$.

For the discussions in this paper, we will use the variable $t$ to denote the current set of tasks or stream, $\mathcal{D}_{train}^t$ will denote the new set of tasks, while $\mathcal{D}_{prev}^t$ will denote the data retained from the previous tasks. i.e, $\mathcal{D}_{prev}^t = \bigcup_{w=1}^{k(t-1)}\{\mathcal{X}_{w_{prev}}^j\}_{j=1}^{m.N}$ where $w$ indexes the previously learned classes, and $m$ is the fraction of data retained. For ease of representation, we may drop the indices to represent the sets of sample and ground truth belonging to $\mathcal{D}_{train}^t$ and $\mathcal{D}_{prev}^t$ as $\{\mathcal{X}_{train}^t, \mathcal{Y}_{train}^t\}$ and $\{\mathcal{X}_{prev}^t, \mathcal{Y}_{prev}^t\}$ respectively.

\subsection{Overview}
We organise the novel classes to be learned based on the curriculum created for each incoming set of tasks $t>1$. The curriculum for a collection of tasks, say $t$, is created using a curriculum generation algorithm based on previously learned information from the old model, $\Theta_{t-1}$, as shown in section~\ref{curriculum-generation}.  Then, when training on the new classes, we use a combination of replay and information distillation techniques to ensure minimal interference with previously learned features.
We also use an insightful subset selection algorithm to prune data in order to make the most of the memory budget, as discussed in section~\ref{ISS}.Algorithm depicting the overall process is added in the supplementary.

\subsection{Curriculum Generation}
\label{curriculum-generation}
In this work we propose a curriculum based continual learning framework building on the idea of learning efficiently using a curriculum based on similarity of concepts. For this we introduce a strategy for devising curriculum based on the similarity between the concepts to be learned. The basic idea is to learn the concepts similar to the previously learned information. We formulate a cosine similarity based feature ranking for the novel set of tasks $t$, with respect to the features learned till the previous set of tasks $t-1$. Solely for the purpose of curriculum generation and to execute equation~\ref{eq:mean-t-1}, we retain the extracted features corresponding to $N$ samples, from each of $k$ classes of the previous set of task $t-1$. Note that no previous data is retained nor these extracted features are used for any sort of training.

Consider the model $\Theta_{t-1}$ trained incrementally on the set of tasks till $t-1$. Now we extract the features  corresponding to the set of tasks $t$ and $t-1$ and obtain the representative prototype per class denoted by $\mu_{l}^q$ for both these set of tasks as shown in Equation~\ref{eq:mean-t-1}.
\begin{equation}
    \mu_{l}^q = \frac{1}{N}\sum_{j=1}^{N} \Theta_{t-1}(x_j^{l,q})
    \label{eq:mean-t-1}
\end{equation}
Where, $l \in \{t-1, t\}$ denoting the previous and current set of tasks, $q$ is used to index the classes in each set of task with each class having a total of $N$ samples.

With these per class prototypes in hand, we now compute the curriculum for set of tasks $t$ as,
\begin{equation}
    C_t = \argmax_{c}  \{[s_{r1}, s_{r2}, \dots, s_{rc}, \dots, s_{rk} ]\}_{r=1}^{k} 
    \label{eq:curric}
\end{equation}
where $s_{rc}$ denotes an element of the similarity matrix $S$, denoting the similarity measure between the class prototypes of the class $r$ from the previous set of task $t-1$ with that of the $k$ classes from the current set of tasks. This similarity matrix $S$ is computed as shown in Equation~\ref{eq:cosine},
\begin{equation}
    s_{rc} =  \frac{\mu_r^T \mu_c}{\left\Vert \mu_r \right\Vert \left\Vert \mu_c \right\Vert}
    \label{eq:cosine}
\end{equation}
The argmax function in Equation~\ref{eq:curric}  gives us the index $r$ indicating closest category to the previous set of tasks $t-1$ among the current set of tasks $t$. Thus a $k$ dimensional vector $C_t$ gives us the curriculum to train the model $\Theta_t$ incrementally as discussed in the section~\ref{cbcl}. 
\subsection{Curriculum based Continual Learning}
\label{cbcl}
The training of the model $\Theta$ can be split into two phases with the initial phase of training, for $t=1$ performed over a random set of $k$ classes which will form the first set of tasks. For this phase we train the model using the traditional multi-class cross entropy loss function as can be seen in Equation~\ref{eq:ce}.

After this initial training phase we have the continual learning phase with the new set of incoming tasks i.e, $t=2, t=3, \dots$ and so on. For each set of incoming task, say $t$ we will generate the curriculum as explained in section~\ref{curriculum-generation}. Based on the curriculum $C_t$ thus obtained we train on the new set of tasks using a combination of multi class cross-entropy and contrastive loss based regularizer imposed over the distilled logits from the teacher model corresponding to previously learned categories as shown in Equation~\ref{eq:total-cost}. The contrastive distillation based regularization, $\mathcal{R}$ is applied only to the output of the old classification layers which corresponds to the previous classes. While the  multi-class  cross-entropy loss, $\mathcal{L_C}$ is  employed  upon all the classification layers corresponding to both old and new classes.
\begin{equation}
     \mathcal{L_C} = -\frac{1}{N} \sum_{i=1}^{N}\sum_{j=1}^{k} p_{ij}\log q_{ij}
     \label{eq:ce}
     \end{equation}
\begin{equation}
    \mathcal{R} = -\frac{1}{N} \sum_{i=1}^{N}\sum_{j=1}^{P}
    \log \frac{\exp( p_{ij}^{\prime}. q_{ij}^\prime )}{\sum_j \exp(p_{ij}^{\prime}. q_{ij}^\prime)}
    \label{eq:dist}
\end{equation}
\begin{equation}
    \mathcal{L}_T = \mathcal{L_C} + \mathcal{R}
    \label{eq:total-cost}
\end{equation}
 $N$ is the number of samples, and $P$ denotes the total number of previous classes. $p_{ij}^{\prime}$ is the ground truth and $q_{ij}^{\prime}$ denotes the softmax output for the $i^{th}$ sample of the $j^{th}$.
For proper transfer of the already learned features of previous tasks we employ the knowledge distillation technique as,
\begin{equation}
    p_{i}^\prime = \frac{\exp{(z_i/T)}}{\sum_j \exp{(z_j/T)}}
    \label{eq:kd}
\end{equation}
where, $p_i$ is obtained by performing distillation over the logit $z_i$ using $T$ which is called the temperature parameter. The value of $T$ is kept equal to $2$ for optimal performance as suggested in the work~\cite{hinton2015distilling}.

This is then followed by a representative subset selection stage as explained in the section~\ref{ISS}. Also every phase of learning continually over the incoming set of new tasks is followed by a fine-tuning stage as followed in~\cite{castro2018end} and likes, employed to remove the bias due to class imbalance. 
\subsection{Informative Subset Selection}
\label{ISS}
After each incremental training stage explained in section~\ref{cbcl}, followed by the fine-tuning stage. We select a pool of $\epsilon \times N$ samples from each category in the current set of tasks for representing the corresponding classes in the subsequent learning stages. We try to incorporate the idea of teacher based sample selection for informative subset selection, goal of which is to maximize the efficiency of the continual learning process using this limited budget of retained samples. 

We start by extracting the features $\bigcup_{i=1}^k\{\Theta_t(\mathcal{X}^{ij})\}_{j=1}^N$ corresponding to the $N$ samples of each class in the current set of novel tasks. The extracted features is then subjected to k-means clustering. We obtain the cluster centroids $\{\Gamma^l\}_{l=1}^k$ for the clusters $\{K_l\}_{l=1}^k$. Now, we compute the probability of each feature in the given class to belong to the corresponding centroid by calculating the feature wise squared distance from the centroid as shown in Equation~\ref{eq:sq-dist-prob}.
\begin{equation}
   p(\Theta_t(\mathcal{X}^{ij}) \in K_l) = \frac{\exp{(-\left \Vert \Theta_t(\mathcal{X}^{ij}) - \Gamma^j \right\Vert_2^2)}}{\sum_{v=1}^k \exp{(-\left \Vert \Theta_t(\mathcal{X}^{ij}) - \Gamma^v \right\Vert_2^2)}}
    \label{eq:sq-dist-prob}
\end{equation}
This is used to get the probability vector $\mathcal{P}$ as shown in Equation~\ref{eq:prob}
\begin{equation}
    \mathcal{P}(\Theta_t(\mathcal{X}^{ij}))=[p(\Theta_t(\mathcal{X}^{ij}) \in K_1), \dots , p(\Theta_t(\mathcal{X}^{ij}) \in K_k)]
    \label{eq:prob}
\end{equation}
Now we compute the entropy of each image with respect to the cluster centroid of their corresponding class as,
\begin{equation}
    \mathcal{H}(\mathcal{P}_{ij}) = - \sum_{i=1}^k  p(\Theta_t(\mathcal{X}^{ij}) \in K_l) \log  (p(\Theta_t(\mathcal{X}^{ij}) \in K_l))
\end{equation}
Based on the computed entropy per sample in each cluster we, select a subset by pruning $N(1-\epsilon)$ samples which is farthest from the cluster center. Thus the informative  subset $\mathcal{S}_t$ of the current set of tasks $t$ can be represented as,
\begin{equation}
    \mathcal{S}_t \subsetneq \bigcup_{i=1}^k\{\Theta_t(\mathcal{X}^{ij})\}_{j=1}^N
\end{equation}
such that, $\left| S_t\right| = \epsilon k N $.
\begin{table}[t]
\centering
\begin{tabular}{@{}cccc@{}}
\toprule
\textbf{}                                     & \multicolumn{3}{c}{\textbf{Dataset}}                         \\ \cmidrule(l){2-4} 
\textbf{Method}                               & {NWPU}  & {EuroSat} & {PatternNet} \\ \midrule
\multicolumn{1}{c|}{{LwF~\cite{li2017learning}}}             & 30.17          & 45.67                 & 29.79               \\
\multicolumn{1}{c|}{{EwC~\cite{kirkpatrick2017overcoming}}}             & 20.50          & 24.00                 & 30.39               \\
\multicolumn{1}{c|}{{E2E~\cite{castro2018end}}}             & 29.40          & 78.69                & 27.08               \\
\multicolumn{1}{c|}{{DR~\cite{hou2018lifelong}}}              & 22.09          & -                     & -                   \\
\multicolumn{1}{c|}{{iCaRL~\cite{rebuffi2017icarl}}}           & 37.20          & 76.81                     & 60.97               \\
\multicolumn{1}{c|}{{RS~\cite{ammour2020continual}}}              & 36.54          & -                     & 57.38               \\
\multicolumn{1}{c|}{{MIR~\cite{NIPS2019_9357}}}             & 27.95          & 53.42                     & 53.82               \\
\multicolumn{1}{c|}{{GDUMB~\cite{prabhu2020greedy}}}           & 39.68          & 77.90                     & 59.02               \\
\multicolumn{1}{c|}{{CN-DPM~\cite{lee2020neural}}}          & 20.58          & 65.47                    & 37.40               \\
\multicolumn{1}{c|}{\textbf{Proposed}}        & \textbf{85.73} & \textbf{83.80}        & \textbf{93.67}      \\ \midrule
\multicolumn{1}{c|}{{w/o curriculum}}  & 75.40          & 76.67                 & 84.75               \\
\multicolumn{1}{c|}{{w/o subset selection}} & 78.90          & 78.41                     & 89.63                   \\ \bottomrule
\end{tabular}
\caption{Comparison of accuracy (in \%) on NWPU-RESISC45, EuroSat and PatternNet dataset as shown, using existing approaches. ('-' denotes the result is not reported)}
\label{table:comparison}
\end{table}

\begin{table*}[]

\begin{adjustbox}{width=\linewidth,center}
\centering
\begin{tabular}{cccccccc}
\toprule
{ }                                  & { \textbf{Joint*}}            & \multicolumn{2}{c}{{ \textbf{Proposed}}}                                         & \multicolumn{2}{c}{{ \textbf{w/o curriculum and ISS}}}                                    & { \textbf{w/o curriculum}}  & { \textbf{w/o ISS}} \\ \cmidrule(l){2-8} 
\multirow{-2}{*}{{ \textbf{Stream}}} & { \textbf{Accuracy}} & { \textbf{Accuracy}} & { \textbf{Forgetting}} & { \textbf{Accuracy}} & { \textbf{Forgetting}} & { \textbf{Accuracy }} & { \textbf{Accuracy}}               \\ \hline
{ \textbf{1}}                        & { }                          & {92.76}                     & { --}                           & {93.14}                     & { --}                           & { 91.62}                     & {81.33}                                   \\
{ \textbf{2}}                        & { }                          & {93.65}                     & {1.52}                       & {69.02}                     & {11.05}                        & {69.90}                     & {76.00}                                   \\
{ \textbf{3}}                        & { }                          & {86.41}                     & {0.21}                       & {78.51}                     & {1.08}                       & {78.35}                     & {89.90}                                   \\
{ \textbf{4}}                        & { }                          & {89.75}                     & { 0.38}                       & {81.16}                     & {1.28}                       & { 82.99}                     & {87.03}                                   \\
{ \textbf{5}}                        & {87.59}                     & {89.10}                     & {1.05}                       & {77.02}                     & {7.906}                       & {78.83}                     & {82.53}                                   \\
{ \textbf{6}}                        & { }                          & {86.25}                     & {1.25}                       & {77.47}                     & {6.703}                       & {78.15}                     & {78.23}                                   \\
{ \textbf{7}}                        & { }                          & {82.90}                     & {1.96}                       & {75.48}                     & {7.602}                       & {73.67}                     & {76.72}                                   \\
{ \textbf{8}}                        & { }                          & {79.30}                     & {4.03}                       & {71.34}                     & {11.02}                       & {72.47}                     & {72.64}                                   \\
{ \textbf{9}}                        & { }                          & {78.48}                     & {2.61}                       & {68.41}                     & {12.75}                       & {68.87}                      & {68.19}                                    \\ \bottomrule
\end{tabular}
\end{adjustbox}
\caption{Ablation results  on the proposed framework in the presence and absence of curriculum and subset selection as obtained for the NWPU-RESISC45 dataset. All the reported accuracy and forgetting values are in \%. ISS stands for Informative Subset Selection. * not a result obtained on incremental learning setting}
\label{tab:cvprw-ablation}
\end{table*}

\section{Experiments}
In this section we discuss about the experiments conducted to evaluate the proposed approach. In the following sections we discuss about the datasets used and also on some selected comparisons. Finally we show some ablation studies and discuss on the same.
\subsection{Datasets and Implementation details}
We show our results mainly on NWPU-RESISC45~\cite{cheng2017remote}, developed by Northwestern Polytechnical University as a publicly accessible benchmark for Remote Sensing Image Scene Classification (RESISC) (NWPU). This dataset contains 31,500 images divided into 45 categories.
Other datasets which we conduct experiments on include PatternNet~\cite{zhou2018patternnet},
a high-resolution remote sensing image dataset collected using Google Earth imagery or the Google Map API for satellite image retrieval, divided into 38 classes and RGB version of EuroSat~\cite{helber2019eurosat} dataset based on Sentinel-2 satellite images containing 27000 labelled and geo-referenced samples divided into ten classes. 
As the pipeline's foundation, we use the pre-trained VGG-16 network. For initialising, we use~\cite{glorot2010understanding} and after each convolutional layer, we use batch-norm~\cite{ioffe2015batch}.

We use the Adam optimizer for optimization, with a learning rate of $1\times10^{-6}$ and a weight decay of $1\times10^{-4}$. 
Each input image is resized to $224\times224$ pixels and the number of training epochs is kept at 40 for each set of tasks. We extract a 128 dimension feature vector, and hence we add a 128-dimension fully connected penultimate layer to the architecture.
The value of $\epsilon$ is set at $0.3$ and we use the Adam optimizer with a learning rate of $1\times10^{-7}$ for fine-tuning for 30 epochs. The experiments were conducted using a single Nvidia GeForce GTX 1080 Ti GPU.

To evaluate the performance of our model, we report the average accuracy starting from the incremental stage, training time taken in hours and also the forgetting measure as described in~\cite{chaudhry2018riemannian}.
\subsection{Comparison with existing works}
The results of the proposed solution were compared to the results of various algorithms in the following experiments:
1. The implementation as done in~\cite{li2017learning} is referred to as LwF.
2. EWC refers to the work done in~\cite{kirkpatrick2017overcoming} with elastic weight consolidation, and 3. E2E refers to the end-to-end incremental learning work proposed in~\cite{castro2018end}. 
4. The work suggested in~\cite{rebuffi2017icarl} is referred to as iCaRL and 5. RS denotes the replay based work in~\cite{ammour2020continual}.
Please notice that for reference, the terms DR, MIR, GDUMB, and CN-DPM are used, which refer to the works proposed in~\cite{hou2018lifelong}, \cite{NIPS2019_9357}, \cite{prabhu2020greedy}, and \cite{lee2020neural}, respectively. 

To reproduce these findings, we depend on the extensive work presented in~\cite{mai2021online} and the code given here\footnote{https://github.com/RaptorMai/online-continual-learning}. The results for EWC are obtained using the code contained at \footnote{https://github.com/xialeiliu/RotateNetworks} while the code from this url \footnote{https://github.com/kibok90/iccv20} is used to get E2E and LwF.

For the NWPU-RESISC45, EuroSat, and PatternNet datasets, the Table~\ref{table:comparison} displays the results for a fixed incremental step size of five classes, two classes, and six classes per set of task, respectively.
Our method outperforms the current algorithms by a margin of around $45\%$ and $18.33\%$ for the NWPU-RESISC45 and EuroSat datasets, respectively, and by $32.7\%$ on the PatternNet dataset, from the closest performing algorithm based on these findings. 
\begin{figure}
  \begin{minipage}[b]{0.49\textwidth}
    \centering
    \includegraphics[width=\linewidth]{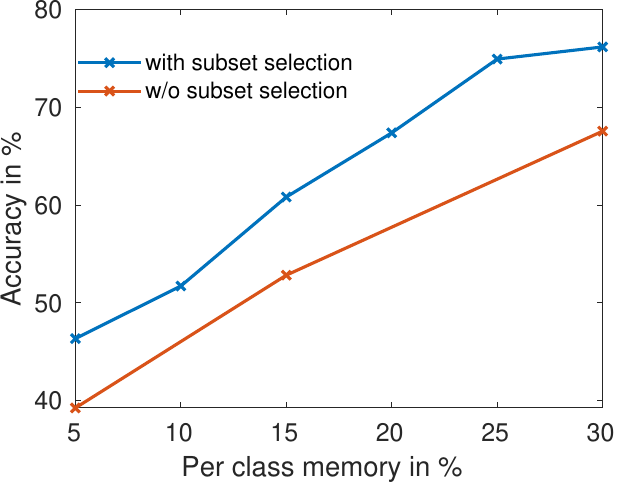}
    \captionof{figure}{Accuracy of different per class memory sizes for NWPU-RESISC45 dataset for a step size of 9 classes.}
        \label{table:mem}
  \end{minipage}
  \hfill
  \begin{minipage}[b]{0.49\textwidth}
    \centering
    \includegraphics[width=\linewidth]{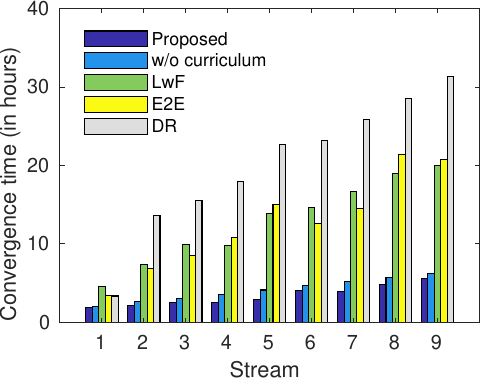}
\caption{ Convergence time Vs stream plot for comparing multiple methods. The result is shown for a step size of 5 on NWPU dataset.}
\label{fig:nwpu-time}
    \end{minipage}
  \end{figure}
\subsection{Ablation Studies}
In this paper, we propose an efficient continual learning system for satellite image classification. By promoting the learning process and assisting it in achieving a better optimum, we expect the curriculum learning approach to reduce the time it takes to converge. The proposed subset selection for sample retention aims to utilize the sample retention budget we have effectively. The suggested solution is examined in the following sections when the curriculum is omitted, subset selection is missing, or a mix of these.

\begin{table}[t]
\centering
\begin{adjustbox}{width=0.7\columnwidth,center}
\begin{tabular}{@{}ccccc@{}}
\toprule
\multirow{2}{*}{Step size} & \multicolumn{2}{c}{Accuracy (in \%)}                                                                                                                             & \multicolumn{2}{c}{Time taken (in hrs)}                                                                                                                          \\ \cmidrule(l){2-5} 
                           & \multicolumn{1}{c}{\begin{tabular}[c]{@{}c@{}}Our \\ Approach\end{tabular}} & \multicolumn{1}{c}{\begin{tabular}[c]{@{}c@{}}w/o \\ curriculum\end{tabular}} & \multicolumn{1}{c}{\begin{tabular}[c]{@{}c@{}}Our \\ Approach\end{tabular}} & \multicolumn{1}{c}{\begin{tabular}[c]{@{}c@{}}w/o \\ curriculum\end{tabular}} \\ \midrule

\multicolumn{1}{c}{5}    & \multicolumn{1}{c}{85.73}                                                      & \multicolumn{1}{c}{75.40}                                                     & \multicolumn{1}{c}{32.05}                                                       & \multicolumn{1}{c}{39.29}                                                      \\ \midrule
\multicolumn{1}{c}{9}    & \multicolumn{1}{c}{76.17}                                                      & \multicolumn{1}{c}{63.31}                                                     & \multicolumn{1}{c}{20.17}                                                       & \multicolumn{1}{c}{30.75}  
                                                                  \\ \bottomrule
\end{tabular}
\end{adjustbox}
\caption{Accuracy and total time consumed for different step sizes for the NWPU-RESISC45 dataset for proposed approach and without a curriculum.}
\label{table:nwpu}
\end{table}

\noindent \textbf{Convergence Time Comparison}
To demonstrate the improved convergence time, we plot the time taken per stream to converge for both the curriculum-based and curriculum-less approaches. The time required is calculated based on the 40 epochs of training and includes the curriculum development and fine-tuning stages for consistency.

In contrast to other methods, it is clear that our system has the fastest convergence time. Compared to DR~\cite{hou2018lifelong}'s the highest time of 182.09 hours and 39.29 hours when the curriculum is excluded from our system, the total time for the whole process is around 32.05 hours. 
These findings are also consistent across different step sizes as presented in Table~\ref{table:nwpu}, where for a step size of 9, the proposed approach is quicker and converges in 20.17 hours while it takes 30.75 hours when the curriculum is removed.

\begin{figure*}
    \centering
    \subfloat[][Accuracy per stream for the NWPU-RESISC45 dataset with a step size of 5.]{\includegraphics[width=0.5\textwidth]{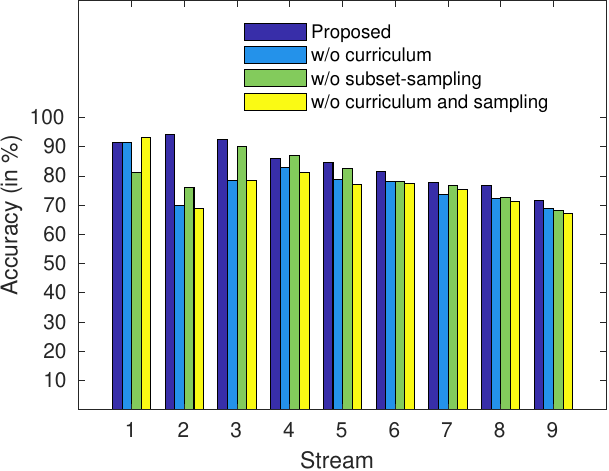}\label{fig:nwpu_step5-acc}}
\hfill
    \subfloat[][Average Forgetting per stream for NWPU-RESISC45 data with a step size of 5.]{\includegraphics[width=0.5\textwidth]{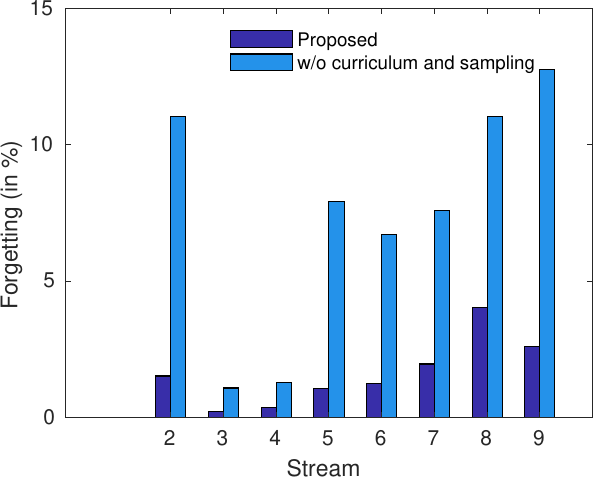}\label{fig:nwpu_step5-forg}}

    \caption{Results for accuracy and forgetting at each stream for the NWPU-RESISC45 dataset. The results are shown for both with and without curriculum approach.}
    
\end{figure*}

\noindent \textbf{Accuracy and extent of forgetting}
Table~\ref{table:nwpu} shows the average accuracy obtained for the proposed framework and curriculum-less learning for different incremental step sizes.
From the Table~\ref{table:comparison} and \ref{tab:cvprw-ablation}, we observe that the proposed framework outperforms those without curriculum by a margin of around $10\%$ and with the subset selection removed by a margin of $7\%$ for the step size considered. Also, it is to be noted that the extent of forgetting is relatively minimal for the proposed approach, as shown in Figure~\ref{fig:nwpu_step5-forg}. We also see that the joint training accuracy is higher, as anticipated by a small margin. However, it should be remembered that this is an outcome of training the network with all classes together in a conventional manner and not how real-world systems operate.

\noindent \textbf{Effect of memory budget}
This section looks at how the model's output changes when the per-class memory budget changes. It is clear from Figure~\ref{table:mem} that the output degrades as the number of samples retained per class is reduced. This pattern is in line with what one would expect as the number of samples used to memorize previous classes decreases. 
However, note that the extent of this impact is mitigated with use of subset selection algorithm.

Nonetheless, for the NWPU dataset, the efficiency decline for the proposed approach with change in per-class memory is by a margin of $29.81\%$ while the amount of decrease in samples retained ranges from $5-30\%$, accounting for a $25\%$ variation in memory usage.

\section{Conclusions}
In this paper, we propose a novel CIL framework for tackling the problem of incremental land-cover classification from optical remote sensing images. Our method is developed on the premise of rehearsal-based memory replay, where a subset of samples from the old classes is preserved and replayed periodically. However, there are two inherent problems of such an approach which we address here. First, we limit the catastrophic forgetting by using curriculum learning on the novel classes by exploiting their similarities with the old classes. Next, we propose an improved replay memory by selecting a highly reliable subset from the old classes. We experimentally find that the proposed modifications offer an improved stability-plasticity trade-off for CIL with respect to a number of baseline techniques from the literature. We are currently interested in extending our model for non-rehearsal-based memory replay models and considering the possibility of having unknown novel classes in each episode.


\bibliography{mybibfile}

\begin{thebibliography}{10}
\expandafter\ifx\csname url\endcsname\relax
  \def\url#1{\texttt{#1}}\fi
\expandafter\ifx\csname urlprefix\endcsname\relax\def\urlprefix{URL }\fi
\expandafter\ifx\csname href\endcsname\relax
  \def\href#1#2{#2} \def\path#1{#1}\fi

\bibitem{BELWARD2015115}
A.~S. Belward, J.~O. Skøien,
  \href{http://www.sciencedirect.com/science/article/pii/S0924271614000720}{Who
  launched what, when and why; trends in global land-cover observation capacity
  from civilian earth observation satellites}, ISPRS Journal of Photogrammetry
  and Remote Sensing 103 (2015) 115 -- 128, global Land Cover Mapping and
  Monitoring.
\newblock \href
  {http://dx.doi.org/https://doi.org/10.1016/j.isprsjprs.2014.03.009}
  {\path{doi:https://doi.org/10.1016/j.isprsjprs.2014.03.009}}.
\newline\urlprefix\url{http://www.sciencedirect.com/science/article/pii/S0924271614000720}

\bibitem{gepperth2016incremental}
A.~Gepperth, B.~Hammer, Incremental learning algorithms and applications, in:
  European symposium on artificial neural networks (ESANN), 2016.

\bibitem{mermillod2013stability}
M.~Mermillod, A.~Bugaiska, P.~Bonin, The stability-plasticity dilemma:
  Investigating the continuum from catastrophic forgetting to age-limited
  learning effects, Frontiers in psychology 4 (2013) 504.

\bibitem{french1999catastrophic}
R.~M. French, Catastrophic forgetting in connectionist networks, Trends in
  cognitive sciences 3~(4) (1999) 128--135.

\bibitem{bengio2009curriculum}
Y.~Bengio, J.~Louradour, R.~Collobert, J.~Weston, Curriculum learning, in:
  Proceedings of the 26th annual international conference on machine learning,
  ACM, 2009, pp. 41--48.

\bibitem{skinner1958reinforcement}
B.~F. Skinner, Reinforcement today., American Psychologist 13~(3) (1958) 94.

\bibitem{krueger2009flexible}
K.~A. Krueger, P.~Dayan, Flexible shaping: How learning in small steps helps,
  Cognition 110~(3) (2009) 380--394.

\bibitem{paul2016efficient}
S.~Paul, J.~H. Bappy, A.~K. Roy-Chowdhury, Efficient selection of informative
  and diverse training samples with applications in scene classification, in:
  2016 IEEE International Conference on Image Processing (ICIP), IEEE, 2016,
  pp. 494--498.

\bibitem{settles2009active}
B.~Settles, Active learning literature survey.

\bibitem{vondrick2011video}
C.~Vondrick, D.~Ramanan, Video annotation and tracking with active learning,
  Advances in Neural Information Processing Systems 24 (2011) 28--36.

\bibitem{vijayanarasimhan2014large}
S.~Vijayanarasimhan, K.~Grauman, Large-scale live active learning: Training
  object detectors with crawled data and crowds, International journal of
  computer vision 108~(1) (2014) 97--114.

\bibitem{rebuffi2017icarl}
S.-A. Rebuffi, A.~Kolesnikov, G.~Sperl, C.~H. Lampert, icarl: Incremental
  classifier and representation learning, in: Proceedings of the IEEE
  conference on Computer Vision and Pattern Recognition, 2017, pp. 2001--2010.

\bibitem{ammour2020continual}
N.~Ammour, Y.~Bazi, H.~Alhichri, N.~Alajlan, Continual learning approach for
  remote sensing scene classification, IEEE Geoscience and Remote Sensing
  Letters.

\bibitem{10.1007/978-3-540-71701-0_101}
Y.-M. Wen, B.-L. Lu, Incremental learning of support vector machines by
  classifier combining, in: Z.-H. Zhou, H.~Li, Q.~Yang (Eds.), Advances in
  Knowledge Discovery and Data Mining, Springer Berlin Heidelberg, Berlin,
  Heidelberg, 2007, pp. 904--911.

\bibitem{bruzzone1999incremental}
L.~Bruzzone, D.~F. Prieto, An incremental-learning neural network for the
  classification of remote-sensing images, Pattern Recognition Letters
  20~(11-13) (1999) 1241--1248.

\bibitem{lakshminarayanan2014mondrian}
B.~Lakshminarayanan, D.~M. Roy, Y.~W. Teh, Mondrian forests: Efficient online
  random forests, in: Advances in neural information processing systems, 2014,
  pp. 3140--3148.

\bibitem{kirkpatrick2017overcoming}
J.~Kirkpatrick, R.~Pascanu, N.~Rabinowitz, J.~Veness, G.~Desjardins, A.~A.
  Rusu, K.~Milan, J.~Quan, T.~Ramalho, A.~Grabska-Barwinska, et~al., Overcoming
  catastrophic forgetting in neural networks, Proceedings of the national
  academy of sciences 114~(13) (2017) 3521--3526.

\bibitem{liu2018rotate}
X.~Liu, M.~Masana, L.~Herranz, J.~Van~de Weijer, A.~M. Lopez, A.~D. Bagdanov,
  Rotate your networks: Better weight consolidation and less catastrophic
  forgetting, in: 2018 24th International Conference on Pattern Recognition
  (ICPR), IEEE, 2018, pp. 2262--2268.

\bibitem{zenke2017continual}
F.~Zenke, B.~Poole, S.~Ganguli, Continual learning through synaptic
  intelligence, Proceedings of machine learning research 70 (2017) 3987.

\bibitem{li2017learning}
Z.~Li, D.~Hoiem, Learning without forgetting, IEEE transactions on pattern
  analysis and machine intelligence 40~(12) (2017) 2935--2947.

\bibitem{hinton2015distilling}
G.~Hinton, O.~Vinyals, J.~Dean, Distilling the knowledge in a neural network,
  arXiv preprint arXiv:1503.02531.

\bibitem{castro2018end}
F.~M. Castro, M.~J. Mar{\'\i}n-Jim{\'e}nez, N.~Guil, C.~Schmid, K.~Alahari,
  End-to-end incremental learning, in: Proceedings of the European Conference
  on Computer Vision (ECCV), 2018, pp. 233--248.

\bibitem{tasar2019incremental}
O.~Tasar, Y.~Tarabalka, P.~Alliez, Incremental learning for semantic
  segmentation of large-scale remote sensing data, IEEE Journal of Selected
  Topics in Applied Earth Observations and Remote Sensing 12~(9) (2019)
  3524--3537.

\bibitem{wu2018incremental}
Y.~Wu, Y.~Chen, L.~Wang, Y.~Ye, Z.~Liu, Y.~Guo, Z.~Zhang, Y.~Fu, Incremental
  classifier learning with generative adversarial networks, arXiv preprint
  arXiv:1802.00853.

\bibitem{NIPS2019_9357}
R.~Aljundi, E.~Belilovsky, T.~Tuytelaars, L.~Charlin, M.~Caccia, M.~Lin,
  L.~Page-Caccia,
  \href{http://papers.nips.cc/paper/9357-online-continual-learning-with-maximal-interfered-retrieval.pdf}{Online
  continual learning with maximal interfered retrieval}, in: H.~Wallach,
  H.~Larochelle, A.~Beygelzimer, F.~d\textquotesingle Alch\'{e}-Buc, E.~Fox,
  R.~Garnett (Eds.), Advances in Neural Information Processing Systems 32,
  Curran Associates, Inc., 2019, pp. 11849--11860.
\newline\urlprefix\url{http://papers.nips.cc/paper/9357-online-continual-learning-with-maximal-interfered-retrieval.pdf}

\bibitem{prabhu2020greedy}
A.~Prabhu, P.~Torr, P.~Dokania, Gdumb: A simple approach that questions our
  progress in continual learning, in: The European Conference on Computer
  Vision (ECCV), 2020.

\bibitem{lee2020neural}
S.~Lee, J.~Ha, D.~Zhang, G.~Kim, A neural dirichlet process mixture model for
  task-free continual learning, arXiv preprint arXiv:2001.00689.

\bibitem{yang2020geoboost}
N.~Yang, H.~Tang, Geoboost: an incremental deep learning approach toward global
  mapping of buildings from vhr remote sensing images, Remote Sensing 12~(11)
  (2020) 1794.

\bibitem{li2020clrs}
H.~Li, H.~Jiang, X.~Gu, J.~Peng, W.~Li, L.~Hong, C.~Tao, Clrs: Continual
  learning benchmark for remote sensing image scene classification, Sensors
  20~(4) (2020) 1226.

\bibitem{hou2018lifelong}
S.~Hou, X.~Pan, C.~C. Loy, Z.~Wang, D.~Lin, Lifelong learning via progressive
  distillation and retrospection, in: Proceedings of the European Conference on
  Computer Vision (ECCV), 2018, pp. 437--452.

\bibitem{cheng2017remote}
G.~Cheng, J.~Han, X.~Lu, Remote sensing image scene classification: Benchmark
  and state of the art, Proceedings of the IEEE 105~(10) (2017) 1865--1883.

\bibitem{zhou2018patternnet}
W.~Zhou, S.~Newsam, C.~Li, Z.~Shao, Patternnet: A benchmark dataset for
  performance evaluation of remote sensing image retrieval, ISPRS journal of
  photogrammetry and remote sensing 145 (2018) 197--209.

\bibitem{helber2019eurosat}
P.~Helber, B.~Bischke, A.~Dengel, D.~Borth, Eurosat: A novel dataset and deep
  learning benchmark for land use and land cover classification, IEEE Journal
  of Selected Topics in Applied Earth Observations and Remote Sensing.

\bibitem{glorot2010understanding}
X.~Glorot, Y.~Bengio, Understanding the difficulty of training deep feedforward
  neural networks, in: Proceedings of the thirteenth international conference
  on artificial intelligence and statistics, 2010, pp. 249--256.

\bibitem{ioffe2015batch}
S.~Ioffe, C.~Szegedy, Batch normalization: Accelerating deep network training
  by reducing internal covariate shift, arXiv preprint arXiv:1502.03167.

\bibitem{chaudhry2018riemannian}
A.~Chaudhry, P.~K. Dokania, T.~Ajanthan, P.~H. Torr, Riemannian walk for
  incremental learning: Understanding forgetting and intransigence, in:
  Proceedings of the European Conference on Computer Vision (ECCV), 2018, pp.
  532--547.

\bibitem{mai2021online}
Z.~Mai, R.~Li, J.~Jeong, D.~Quispe, H.~Kim, S.~Sanner, Online continual
  learning in image classification: An empirical survey, arXiv preprint
  arXiv:2101.10423.

\end{thebibliography}

\end{document}